%% file: main.tex
\documentclass[runningheads]{llncs}

\usepackage{eccv}
\usepackage{eccvabbrv}
\usepackage{graphicx}
\usepackage{booktabs}
\usepackage[accsupp]{axessibility}

\usepackage{hyperref}
\usepackage{orcidlink}

\usepackage{adjustbox}
\usepackage{pifont}
\usepackage{tabularx}
\usepackage{amsmath}
\usepackage{amssymb}
\usepackage{pifont}
\newcommand{\cmark}{\ding{51}}

\usepackage[table]{xcolor}
\usepackage{float}
\usepackage{bbm}
\usepackage{multirow}
\usepackage{wrapfig}
\usepackage{subcaption}
\usepackage{algorithm}
\usepackage{algpseudocode}

\begin{document}

\title{Context Blindness in DPO:\\Mitigating Object Hallucination in MLLMs\\via Context-Calibrated Preference Optimization}
\titlerunning{C$^2$-DPO: Context-Calibrated Preference Optimization}

\author{
    Byungoh Ko\inst{1}\orcidlink{0009-0004-0616-7343} \and 
    Jinyoung Park\inst{2} \and
    Jongha Kim\inst{2} \and 
    Jeehye Na\inst{2} \and
    Jaewon Cho\inst{2} \and
    Hyunwoo J. Kim\inst{2}\orcidlink{0000-0002-2181-9264}\thanks{Corresponding author}
}
\authorrunning{B. Ko et al.}

\institute{
    Korea University, Seoul, Republic of Korea \\
    \email{ko990128@korea.ac.kr} \and
    KAIST, Daejeon, Republic of Korea \\
    \email{hyunwoojkim@kaist.ac.kr}
}

\maketitle

\input{sec/0_abstract}    
\input{sec/1_intro}
\input{sec/2_relwork}
\input{sec/3_preliminaries}
\input{sec/4_method}
\input{sec/5_experiments}
\input{sec/6_conclusion}

\bibliographystyle{splncs04}
\bibliography{main}
\newpage
\input{sec/X_suppl}
\end{document}

%% file: sec/0_abstract.tex
\begin{abstract}
Multimodal large language models (MLLMs) have made rapid progress, yet they still exhibit object hallucination, generating plausible but incorrect descriptions that are inconsistent with the visual input.
Direct Preference Optimization (DPO) mitigates this by training models to prefer non-hallucinated responses over hallucinated ones, and recent efforts further enrich the preference data with relevant context.
However, it remains unclear whether DPO actually leverages such context.
To investigate this, we propose Contextual Preference Gain (CPG), a simple metric that measures how much a model's preference strengthens when relevant context is provided.
We find that higher CPG consistently corresponds to lower hallucination, yet standard DPO and its variants exhibit only limited CPG, indicating that they underutilize contextual information and thus remain prone to hallucination.
To address this, we propose Context-Calibrated DPO (C$^2$-DPO), which directly maximizes CPG while preserving the original preference ordering.
Across multiple benchmarks, C$^2$-DPO substantially reduces hallucination without compromising general reasoning, relatively reducing the Object HalBench hallucination rate of Qwen2-VL-Instruct-2B by 36\%.
Code is available at \url{https://github.com/mlvlab/C2-DPO}
\keywords{Multimodal Large Language Models \and Object Hallucination \and Preference Optimization}
\end{abstract}

%% file: sec/1_intro.tex
\section{Introduction}
\input{figure/fig_example}
\label{sec:intro}

Large language models (LLMs)~\cite{openai2024gpt4o,anthropic2025claude37sonnet,deepmind2025gemini25,meta2025llama4,yang2025qwen3} have shown remarkable performance across a wide range of language tasks. 
Building on these advances, multimodal large language models (MLLMs)~\cite{liu2023visual,dai2023instructblip,wang2024qwen2,bai2023qwen,li2024llava,liu2024improved} integrate visual encoders and multimodal training data, enabling unified understanding of images and text.
Recent MLLMs achieve impressive results on core vision-language tasks such as image captioning~\cite{lin2014microsoft,agrawal2019nocaps} and visual question-answering~\cite{antol2015vqa,goyal2017making,singh2019towards,lu2022learn,yu2023mm}.

Despite the progress, MLLMs remain prone to object hallucination, where the model generates semantically plausible descriptions that are not grounded in the visual input.
This issue undermines the factual reliability of multimodal reasoning.
To mitigate object hallucination, recent works have applied preference optimization, which trains the model to prefer non-hallucinated responses rather than hallucinated responses.
Early studies~\cite{zhao2023beyond,zhou2024aligning,ouali2024clip} primarily focused on constructing high-quality preference data for preference optimization~\cite{rafailov2023direct}.
More recently, a follow-up study~\cite{peng2025mitigating} enriches the input with non-hallucinated descriptions to strengthen contextual grounding, thereby making preferred and dispreferred responses more discriminative and improving alignment performance.

While these approaches modify the preference dataset for better optimization, a more fundamental question remains unanswered: ``Does DPO encourage a model to leverage relevant context and learn contextual preference?''
Motivated by the information-theoretic view that relevant information reduces uncertainty~\cite{burgin2002essence, soni2017mind}, 
we argue that effective preference optimization should strengthen the model's preference for the correct response when richer contextual information is available.
Structurally, DPO optimizes the preference margin for a single fixed input, so its gradient never rewards a larger margin when richer context is supplied.
To investigate this, we introduce Contextual Preference Gain~(CPG), a metric that measures how much the model's preference strengthens when an additional helpful context is provided.
Using CPG, we identify two notable observations.
First, CPG is strongly and negatively correlated with hallucination rate: models with higher CPG consistently achieve lower hallucination rates.
Second, standard DPO and its variants show CPG distributions concentrated near zero, suggesting that their learned preferences are \emph{context-blind}.
Together, these findings indicate that existing preference optimization approaches inadequately exploit contextual information for grounding.

To address this limitation, we propose Context-Calibrated Direct Preference Optimization (C$^2$-DPO), a context-aware preference alignment framework designed to amplify grounding signals for mitigating object hallucination in MLLMs.
C$^2$-DPO explicitly models how helpful context influences the relative preference between preferred and dispreferred responses, and encourages higher preference scores of the input with the additional relevant context.
Specifically, a calibration term directly rewards the preference margin obtained \emph{with} context over the one obtained without it, turning contextual grounding into an explicit training signal rather than an emergent side effect.
By calibrating preference scores with respect to contextual information, C$^2$-DPO enables MLLMs to better leverage input context and reduce object hallucination.

Comprehensive experiments across diverse benchmarks demonstrate that C$^2$-DPO effectively reduces object hallucinations while preserving general reasoning and instruction-following capabilities.
On Object Halbench~\cite{rohrbach2018object}, hallucination rates decrease by 36\% and 60\% in response and mention level, respectively, with consistent improvements observed on HallusionBench~\cite{guan2024hallusionbench}.
Furthermore, C$^2$-DPO maintains its general reasoning performance on ScienceQA, MM-Vet, and TextVQA~\cite{lu2022learn,yu2023mm,singh2019towards}, indicating that mitigating object hallucination does not come at the cost of overall downstream task performance.
Beyond the standard DPO, we further show that our context-aware preference gain formulation can be seamlessly incorporated into other preference-optimization variants such as SimPO~\cite{meng2024simpo} and RDPO~\cite{park2024disentangling}, consistently yielding similar gains.
Additionally, experiments in text-only settings confirm that C$^2$-DPO enhances factual alignment even without visual inputs, suggesting that context-aware preference modeling offers broad benefit beyond multimodal domains.

Our contributions are summarized as follows:
\begin{itemize}
    \item We introduce Contextual Preference Gain (CPG), a diagnostic metric that quantifies how preference margins change with additional grounding information, and reveal a strong correlation between CPG and hallucination rate, exposing a \textit{context blindness} phenomenon in existing DPO-style objectives.
    \item We propose C$^2$-DPO, a general preference alignment framework that explicitly optimizes contextual preference gain and integrates seamlessly with various preference optimization methods such as DPO, SimPO, and RDPO.
    \item We demonstrate the effectiveness and generality of C$^2$-DPO through extensive experiments across multimodal and text-only benchmarks, achieving substantial reduction in object hallucination while preserving the general reasoning performance.
\end{itemize}

%% file: figure/fig_example.tex
\begin{figure}[t!]
    \centering
    \includegraphics[width=1.0\linewidth]{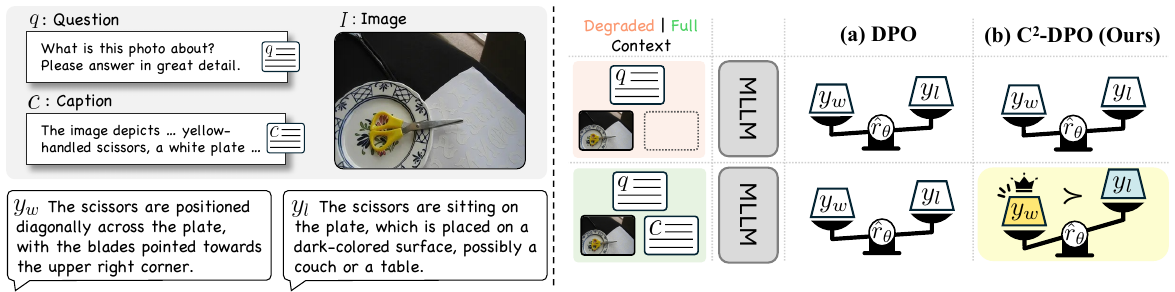}
    \caption{
        \textbf{Motivation of C$^2$-DPO.}
        A model should yield higher preference scores when richer context is provided.
        However, standard DPO shows minimal change between full and degraded contexts, revealing \emph{context-blind} behavior.
        C$^2$-DPO amplifies this gap, enabling models to better leverage contextual information and reduce hallucination.
    }
     
    \label{fig:example}
\end{figure}

%% file: sec/2_relwork.tex
\section{Related Works}
\label{sec:relwork}
\subsection{Hallucination in MLLMs}
Multimodal large language models (MLLMs) often suffer from the ``object hallucination problem'', where generated responses appear natural but deviate from input conditions~\cite{rohrbach2018object,zhou2024analyzing}.
To mitigate this, several inference-time approaches based on Contrastive Decoding (CD)~\cite{li2023contrastive} have been proposed~\cite{leng2024mitigating,huang2024opera,chuang2024dola,chen2024halc}, controlling the model behavior during generation by contrasting multiple outputs.
Still, CD-based approaches remain post-hoc remedies that fail to fundamentally correct the model’s intrinsic behavior, yielding unsatisfactory performance.
Also, they require additional inference cost as multiple forwardings are required during generation.
In parallel, Preference Optimization (PO) methods aim to directly train MLLMs to be less hallucinatory~\cite{peng2025mitigating,zhao2023beyond,zhou2024aligning,yu2025rlaif,ouali2024clip,he2024topic}.
However, prior works focus on constructing better preference datasets, overlooking context blindness, a key limitation that prevents MLLMs from fully leveraging multimodal cues.
Crucially, these methods alter \emph{what} data the model sees but leave the optimization objective unchanged, so the enriched context is not guaranteed to be exploited during alignment.
We identify this issue and propose a solution that further enhances the alignment with the same training data.

\subsection{Direct Preference Optimization (DPO)}
Reinforcement Learning from Human Feedback (RLHF) aligns model behavior with human preferences~\cite{christiano2017deep,ouyang2022training,stiennon2020learning,schulman2017proximal}.
By optimizing models toward preference data, RLHF has proven effective in reducing hallucination in MLLMs~\cite{peng2025mitigating,yu2025rlaif,zhao2023beyond,zhou2024aligning,ouali2024clip,he2024topic}.
Among its variants, Direct Preference Optimization (DPO)~\cite{rafailov2023direct} has gained significant attention for its simplicity, removing the need for an explicit reward model.
However, recent studies have revealed several limitations of DPO, including length bias~\cite{meng2024simpo}, suboptimality of reference models~\cite{xu2024contrastive,meng2024simpo,wualphadpo}, overfitting~\cite{azar2024general}, and inaccurate reward estimation~\cite{xiao2024cal}.
These efforts refine how preference is modeled for a \emph{fixed} input, \eg, by reshaping the reward margin or removing the reference model.
Far less attention has been paid to how the preference should respond when the input context itself varies, which is precisely the regime where multimodal grounding matters.
In this work, we highlight another overlooked issue, \emph{context blindness}, where MLLMs trained with existing preference objectives fail to fully utilize or comprehend input contexts.
To address this, we introduce a new DPO-based training objective that explicitly strengthens contextual understanding in multimodal alignment, offering a principled approach to further mitigate hallucination, complementary to dataset-scaling strategies.

%% file: sec/3_preliminaries.tex
\section{Preliminaries}
\label{sec:preliminaries}
We briefly summarize Direct Preference Optimization for mitigating object hallucination in multimodal large language models~(MLLMs).

Object hallucination in MLLMs refers to the generation of text that does not align with the visual content of the input image.
To address this, preference alignment such as Direct Preference Optimization (DPO)~\cite{rafailov2023direct} has emerged as a promising approach for mitigating the object hallucination problem.
The preference alignment method is designed to train a policy model $\pi_\theta$ to reflect human preferences over candidate responses.
Given an input $x$ and a pair of responses $(y_w, y_l)$ where $y_w$ is the preferred sample while $y_l$ represents the dispreferred sample, DPO optimizes the model to assign higher preference to $y_w$ than $y_l$.

One of the representatives of preference modeling is the Bradley–Terry (BT) framework~\cite{bradley1952rank}, which formulates pairwise preferences as probabilistic outcomes based on differences in underlying rewards.
In the BT model, the probability of preferring $y_w$ over $y_l$ under context $x$ is defined as
\begin{equation}
p\left(y_w \succ y_l \mid x\right) = \sigma\left(r \left(x, y_w \right) - r \left(x, y_l\right)\right),
\end{equation}
where $\sigma(\cdot)$ denotes the sigmoid function and $r \left(x, y \right)$ is the reward associated with response $y$ under context $x$.

Unlike the BT model, which requires an explicit reward function, DPO introduces an implicit reward~$\hat{r}_\theta$ parameterized by the policy model itself:
\begin{equation}
\label{eq:implicit_reward}
\hat{r}_\theta \left(x, y \right) = \beta \log \frac{\pi_\theta \left(y \mid x \right)}{\pi_{\text{ref}} \left(y \mid x \right)},
\end{equation}
where $\pi_\theta$ is the trainable policy model, $\pi_{\text{ref}}$ is a frozen reference model, and $\beta > 0$ controls the regularization strength.
Substituting this implicit reward into the BT formulation yields the DPO loss:
\begin{equation}
\begin{split}
\label{eq:dpo_loss}
    \mathcal{L}_{\text{DPO}}(x, y_w, y_l) 
        &= -\log p(y_w \succ y_l \mid x) \\
        &= -\log \sigma\left(\hat{r}_\theta\left(x, y_w\right) - \hat{r}_\theta\left(x, y_l\right) \right) \\
        &= -\log \sigma \left(
            \beta\log\frac{\pi_\theta(y_w \mid x)}{\pi_\text{ref}(y_w \mid x)} - 
            \beta\log\frac{\pi_\theta(y_l \mid x)}{\pi_\text{ref}(y_l \mid x)}
        \right).
\end{split}
\end{equation}
Thus, DPO can be viewed as maximizing the probability that the model's pairwise preferences align with human judgments, while avoiding the need for an explicit reward model.

Following other DPO works~\cite{peng2025mitigating,zhao2023beyond,zhou2024aligning,yu2025rlaif,ouali2024clip,he2024topic} for mitigating object hallucination, we consider $x = \left(v, q, c \right)$ as the input, where $v$ denotes the input image, $q$ is the text query, and $c$ represents the auxiliary image description helpful for understanding the input image~(\textit{e.g.,} caption).
Here, we treat a non-hallucinated response as the preferred response $y_w$ and a hallucinated response as the dispreferred response $y_l$.
Notably, the DPO objective in Eq.~\eqref{eq:dpo_loss} treats $x$ as a monolithic input and is agnostic to the role of $c$: it never explicitly requires that providing $c$ alter the preference between $y_w$ and $y_l$.
This gap is precisely what we analyze in Section~\ref{subsec:context_blindness}.
For simplicity, since each training instance always includes $(y_w, y_l)$, we denote the DPO loss as $\mathcal{L}_{\text{DPO}}(x)$ hereafter.

%% file: sec/4_method.tex
\section{Method}
\label{sec:main_method}
\subsection{Overview}
Our work aims to optimize multimodal large language models (MLLMs), via preference optimization, to generate outputs that align more closely with the input context.
Most existing approaches~\cite{peng2025mitigating,zhao2023beyond,zhou2024aligning,yu2025rlaif,ouali2024clip,he2024topic} modify the preference dataset, either by rewriting sentences with an external model or by appending enriched image descriptions to the input prompt.
Such modifications make the contrast between hallucinated and non-hallucinated objects more explicit, helping the model understand the preference gap.

However, recent approaches pay little attention to whether the model genuinely grounds this preference gap between preferred~(non-hallucinated) and dispreferred~(hallucinated) samples, leveraging the input context such as images or captions.
We hypothesize that effective preference optimization should lead MLLMs to increasingly favor preferred response $y_w$ over dispreferred response $y_l$ as additional relevant context information is richer.
In other words, the preference gap between $y_w$ and $y_l$ becomes larger as richer contexts~(\eg, captions) are given, while maintaining the preference ranking between them.
This perspective suggests that the model should not only distinguish responses, but also calibrate the strength of this distinction according to the richness of context.
With the preliminary experiments, we found that the standard DPO fails to satisfy these properties, revealing the \emph{context blindness}.

To address this challenge, we present Context-Calibrated Direct Preference Optimization~(C$^2$-DPO), a preference optimization method with a simple calibration term.
Our calibration encourages the model to prefer responses grounded in rich context over degraded context by maximizing the contextual preference gain through a contrastive learning objective.
This lets multimodal large language models better leverage input context during preference alignment, thereby mitigating object hallucination.

\input{figure/fig_correlation}
\subsection{Observations}
\label{subsec:observation}
To take a deep dive into the preference alignment of the standard preference optimization, we measure the Contextual Preference Gain (CPG) according to the richness of the input.
We first introduce the preference score as a measure of alignment strength via implicit reward margin between a preferred response and a dispreferred response:
\begin{equation}
\label{eq:pref_score}
    \Delta \hat{r}_\theta(x, y_w, y_l)
    = \hat{r}_\theta(x, y_w) - \hat{r}_\theta(x, y_l),
\end{equation}
which represents how strongly the model prefers the non-hallucinated response $y_w$ over the hallucinated one $y_l$ given input $x$.

\input{figure/fig_cpg_distribution}  
\input{figure/fig_main_method}

\noindent
\textbf{Contextual Preference Gain (CPG).}
Intuitively, providing additional relevant context should help a model more confidently distinguish a preferred response from a dispreferred one.
As a result, the model should exhibit a stronger preference for the correct and grounded response when richer contextual information is available.
This intuition aligns with the information-theoretic view that relevant information reduces uncertainty~\cite{burgin2002essence, soni2017mind}.
Motivated by this idea, we define Contextual Preference Gain (CPG) as the change in preference score when context is enriched.
Given an input $x$ and its degraded counterpart $x'$, CPG is defined as
\begin{align}
    &\mathrm{CPG}(x, x') =
    \Delta \hat{r}_\theta(x, y_w, y_l) -
    \Delta \hat{r}_\theta(x', y_w, y_l).
    \label{eq:cpg}
\end{align}
A positive CPG indicates that richer or more reliable input leads to a stronger preference for $y_w$ over $y_l$.

\label{subsec:context_blindness}
We evaluate CPG under a context-removal setting where $x'= (v, q, \varnothing)$ corresponds to the input with the auxiliary caption $c$ removed.
As shown in Figure~\ref{fig:correlation}, CPG exhibits a strong negative correlation with hallucination rates on Object HalBench~\cite{rohrbach2018object} and AMBER~\cite{wang2023amber}, where models with higher CPG consistently achieve lower hallucination rates.
This suggests that CPG captures a model's ability to leverage contextual information for grounding.
However, as shown in Figure~\ref{fig:figure2}, existing preference optimization methods~(DPO, SimPO, and RDPO~\cite{rafailov2023direct,meng2024simpo,park2024disentangling}) show CPG distributions highly concentrated near zero, indicating that their learned preferences are largely insensitive to contextual information.
Given the strong association between CPG and hallucination rate, this \emph{context-blind} behavior reveals a fundamental limitation in existing preference optimization methods.

\subsection{Contextual Preference Calibration}
\label{subsec:cpc}
Section~\ref{subsec:context_blindness} shows that the CPG indicates how well a model grounds its responses in context, and thus how prone it is to hallucinate.
Standard preference optimization leaves the CPG concentrated near zero, a failure we term \emph{context blindness}.
A context-calibrated model should instead satisfy the ordering
\begin{equation}
\label{eq:target_ordering}
    \underbrace{\Delta \hat{r}_\theta(x, y_w, y_l)}_{\text{full context}}
    \;>\;
    \underbrace{\Delta \hat{r}_\theta(x', y_w, y_l)}_{\text{degraded context}}
    \;>\; 0,
\end{equation}
where each term is the preference score from Eq.~\eqref{eq:pref_score} under the corresponding context.
The first inequality requires the preference score under the full context to exceed the degraded-context one, which is exactly a positive CPG.
The second requires it to stay positive on its own, so that $y_w$ remains favored over $y_l$ even without the auxiliary cues.
Together, these require the model to preserve the preference ranking across contexts while maximizing the gap as the context becomes richer.

We formulate these desiderata as a primary objective subject to two constraints, taking the full-context DPO loss as the objective and the two inequalities of Eq.~\eqref{eq:target_ordering} as the constraints:
\begin{equation}
\label{eq:constrained}
    \min_{\theta}\ \mathcal{L}_{\text{DPO}}(x)
    \quad\text{s.t.}\quad
    \mathrm{CPG}(x,x') \ge 0,\qquad
    \Delta\hat{r}_\theta(x',y_w,y_l) \ge 0 .
\end{equation}
Rather than solving Eq.~\eqref{eq:constrained} directly, we relax each constraint into a differentiable surrogate that decreases monotonically with its margin, scaled by coefficients $\lambda_c,\lambda_u>0$.
We derive the two surrogates below.

\noindent\textbf{Contextual Preference Calibration Loss.}
For the first constraint, we contrast the two preference scores with a binary NCE-style objective, treating the full context $x$ as the positive and the degraded context $x'$ as the negative:
\begin{align}
\label{eq:cpc_loss}
    \mathcal{L}_c(x,x') 
    &= - \log \frac{\exp\left(\Delta \hat{r}_\theta\left(x,y_w,y_l\right)\right)}
    {\exp\left(\Delta \hat{r}_\theta \left(x,y_w,y_l\right)\right) + \exp(\Delta \hat{r}_\theta(x',y_w,y_l))} \nonumber \\
    &= - \log \sigma \left(
    \Delta \hat{r}_\theta\left(x,y_w,y_l\right) - \Delta \hat{r}_\theta\left(x',y_w,y_l\right)
    \right) \nonumber \\
    &= -\log\sigma\left(\mathrm{CPG}(x,x')\right).
\end{align}
Minimizing $\mathcal{L}_c$ monotonically increases the CPG, driving the full-context preference score above the degraded-context one.
Since a grounded response is favored more strongly as the context grows richer, the model is encouraged to leverage the additional context rather than ignore it.

\noindent\textbf{Preference under the degraded context.}
For the second constraint, we apply the DPO objective to the degraded context $x'$:
\begin{equation}
\label{eq:dpo_xprime}
    \mathcal{L}_{\text{DPO}}(x') = -\log\sigma\!\left(\Delta\hat{r}_\theta(x',y_w,y_l)\right),
\end{equation}
a monotone surrogate that pushes the degraded-context score positive, keeping $y_w$ favored even when the auxiliary cues are removed.
Anchoring this score prevents $\mathcal{L}_c$ from degenerately widening the CPG by lowering $\Delta\hat{r}_\theta(x',y_w,y_l)$ rather than improving grounding.

\noindent\textbf{Training objective.}
Combining the primary objective in Eq.~\eqref{eq:constrained} with the two surrogate terms gives our full training objective:
\begin{equation}
    \mathcal{L}_{\text{C$^2$-DPO}}(x,x') = \mathcal{L}_{\text{DPO}}(x) + \lambda_c\,\mathcal{L}_c(x,x') + \lambda_u\,\mathcal{L}_{\text{DPO}}(x'),
    \label{eq:main}
\end{equation}
where $\lambda_c,\lambda_u>0$ weight the two surrogates against the full-context DPO loss.
Jointly optimizing the three terms encourages the model to satisfy the ordering in Eq.~\eqref{eq:target_ordering}, thereby increasing CPG and reducing object hallucination.
This allows MLLMs to better leverage the input context, improving robustness to hallucination while maintaining a stable preference across input conditions.

%% file: figure/fig_correlation.tex
\begin{figure*}[t]
    \centering
    \includegraphics[width=1.0\textwidth]{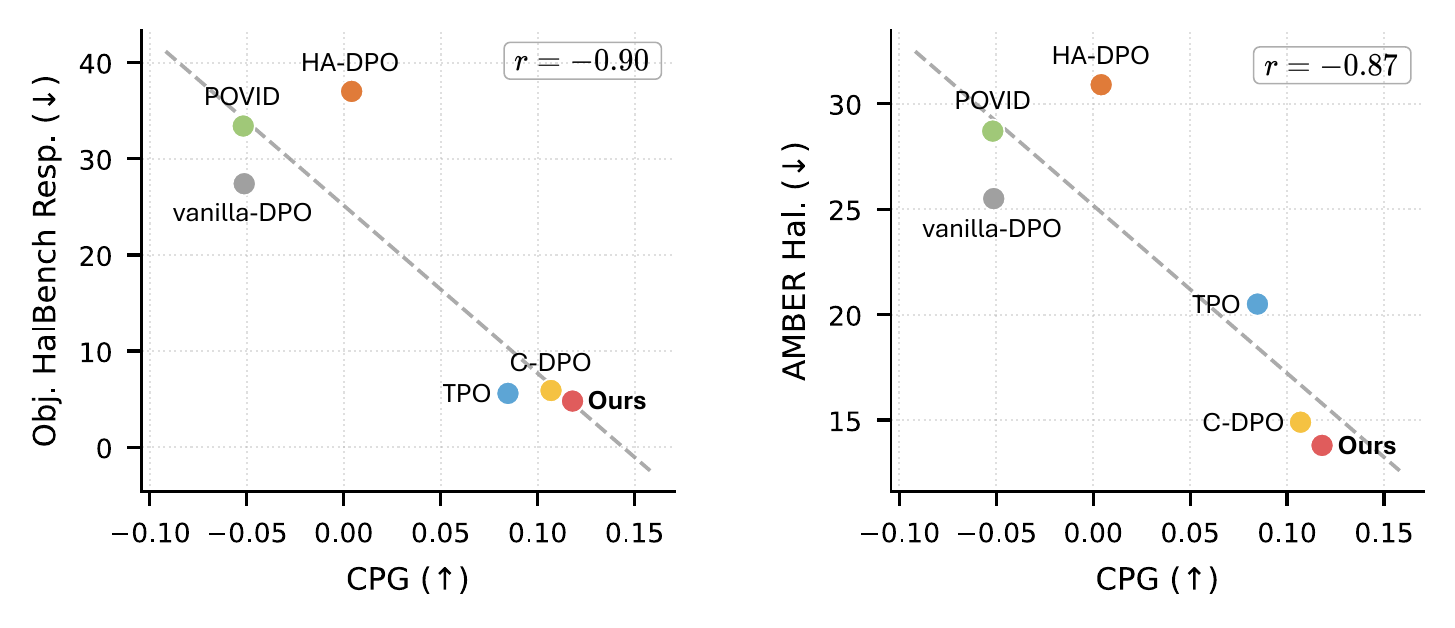}
    
    \caption{
        \textbf{Correlation between CPG and hallucination rates.}
        We analyze the relationship between Contextual Preference Gain (CPG) and hallucination rate across multiple models. 
        On both Object HalBench (left) and AMBER (right), CPG shows a strong negative correlation with hallucination rate, suggesting that increased CPG is closely associated with improved grounding.
    }
    \label{fig:correlation}
\end{figure*}

%% file: figure/fig_cpg_distribution.tex
\begin{figure}[t!]
    \centering
 
    \begin{subfigure}[b]{0.32\linewidth}
        \centering
        \includegraphics[width=\linewidth]{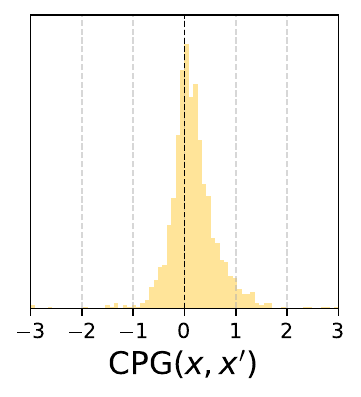}
        \caption{DPO}
    \end{subfigure}\hfill
    \begin{subfigure}[b]{0.32\linewidth}
        \centering
        \includegraphics[width=\linewidth]{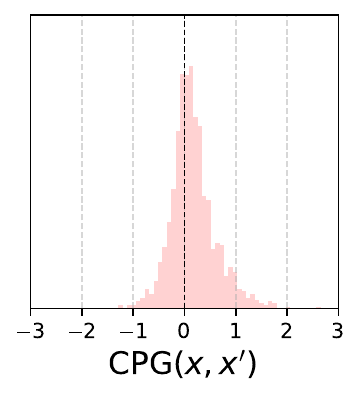}
        \caption{SimPO}
    \end{subfigure}\hfill
    \begin{subfigure}[b]{0.32\linewidth}
        \centering
        \includegraphics[width=\linewidth]{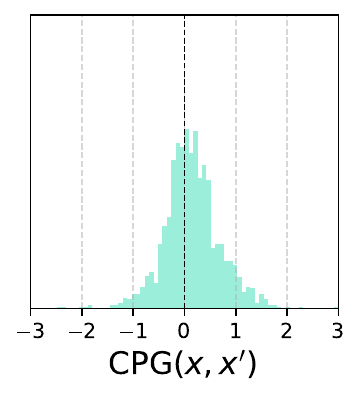}
        \caption{RDPO}
    \end{subfigure}
 
    \caption{
        \textbf{Distributions of Contextual Preference Gain (CPG) for three preference optimization methods: DPO, SimPO, and RDPO.}
        In all cases, the CPG values are highly concentrated near zero, indicating that removing contextual information leads to only minimal changes in preference scores. 
        This demonstrates that existing preference optimization methods are largely \emph{context-blind}.
    }
    \label{fig:figure2}
    
\end{figure}

%% file: figure/fig_main_method.tex
\begin{figure*}[t]
    \centering
    \includegraphics[width=1.0\textwidth]{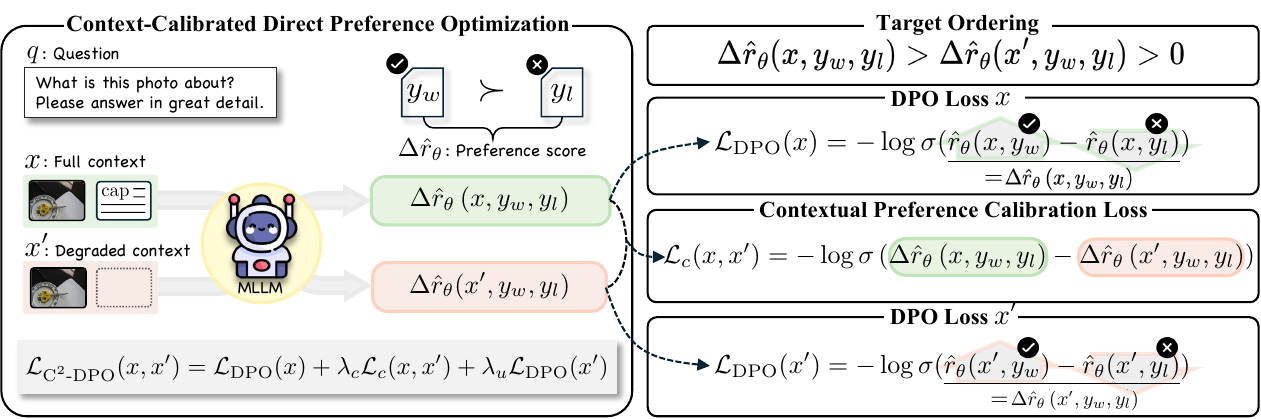}
    \caption{
        \textbf{Context-Calibrated Direct Preference Optimization via Contextual Preference Calibration (C$^2$-DPO).}
        Given a full context $x$ and a degraded context $x'$, C$^2$-DPO calibrates the preference score by (1) applying the standard DPO loss on $x$, (2) contrasting preference scores between $x$ and $x'$ via a Contextual Preference Calibration loss to maximize contextual preference gain, and (3) applying DPO on $x'$ to preserve correct preference ordering under degraded context.
        This joint objective encourages the model to strengthen preference for grounded responses when richer context is available, effectively reducing hallucination.
    }
    \label{fig:main_method}
\end{figure*}

%% file: sec/5_experiments.tex
\section{Experiments}
In this section, we empirically investigate the effectiveness of C$^2$-DPO in enhancing context awareness and reducing hallucinations.
We introduce the experimental setup in Sec.~\ref{subsec:exp_setup}, present the main results in Sec.~\ref{subsec:main_results}, conduct ablation studies in Sec.~\ref{subsec:ablations}, and provide a qualitative analysis in Sec.~\ref{subsec:qual}.

\subsection{Experimental Setup}
\label{subsec:exp_setup}
\textbf{Models and training settings.}
We evaluate our method on two representative MLLMs of different scales: LLaVA-v1.5-7B~\cite{liu2024improved} and Qwen2-VL-Instruct-2B~\cite{wang2024qwen2}.
For a fair comparison, we adopt the same training data as C-DPO~\cite{peng2025mitigating},
using the datasets released for LLaVA-v1.5-7B\footnote{\url{https://huggingface.co/datasets/psp-dada/SENTINEL/blob/main/LLaVA_v1_5_7b_SENTINEL_8_6k.json}}
and Qwen2-VL-Instruct-2B\footnote{\url{https://huggingface.co/datasets/psp-dada/SENTINEL/blob/main/Qwen2_VL_2B_Instruct_SENTINEL_12k.json}}, respectively.
Each preference data consists of an image, a query, and an auxiliary caption, paired with a non-hallucinated (preferred) and a hallucinated (dispreferred) response.
For the degraded input $x'$, we remove the auxiliary caption while keeping the image and query, following the setting in Section~\ref{subsec:context_blindness}; results under alternative degradations (\eg, noisy images, randomly substituted captions) are reported in the supplementary material.
We set $\beta = 0.1$, with $(\lambda_c, \lambda_u)$ set to $(0.3, 0.5)$ for LLaVA-v1.5-7B and $(0.5, 0.3)$ for Qwen2-VL-Instruct-2B, selected from the $[0.3,0.5]$ range based on the sensitivity analysis in Section~\ref{subsec:ablations}.
All models are trained for one epoch using LoRA~\cite{hu2022lora} with AdamW~\cite{loshchilov2017decoupled} at a learning rate of $2 \times 10^{-6}$.
Training is conducted with a global batch size of 64.

\noindent\textbf{Evaluation benchmarks.}
We evaluate the hallucination rates and general multimodal reasoning ability of our C$^2$-DPO approach across a wide range of benchmarks.
To evaluate hallucination, we use Object HalBench~\cite{rohrbach2018object}, AMBER~\cite{wang2023amber}, and HallusionBench~\cite{guan2024hallusionbench}.
To assess general multimodal reasoning performance, we further evaluate on ScienceQA~\cite{lu2022learn}, MM-Vet~\cite{yu2023mm}, and TextVQA~\cite{singh2019towards}.

\noindent\textbf{Baselines.}
We compare C$^2$-DPO with a wide range of state-of-the-art methods.
Specifically, VCD~\cite{leng2024mitigating}, OPERA~\cite{huang2024opera}, and DoLa~\cite{chuang2024dola} apply contrastive decoding strategies, whereas HA-DPO~\cite{zhao2023beyond}, POVID~\cite{zhou2024aligning}, CLIP-DPO~\cite{ouali2024clip}, RLAIF-V~\cite{yu2025rlaif}, TPO~\cite{he2024topic}, and C-DPO~\cite{peng2025mitigating} adopt preference optimization approahces.
We additionally include a vanilla-DPO baseline, obtained by training with the original DPO~\cite{rafailov2023direct} objective on the C-DPO dataset~\cite{peng2025mitigating}, without using any auxiliary captions.
Further details on training, datasets, and evaluation protocols are provided in the supplementary material.

\subsection{Main Results}

\input{table/main_table}
\input{table/ablation_loss}
\noindent\textbf{Comparison with baselines.}
As shown in Table~\ref{tab:main_table}, C$^2$-DPO delivers consistent and substantial improvements over prior methods across a wide range of hallucination benchmarks.
On Object HalBench, our approach achieves the strongest performance among all compared methods.
For Qwen2-VL-Instruct-2B, C$^2$-DPO reduces response-level and mention-level hallucinations to 1.6 and 1.0, outperforming C-DPO~\cite{peng2025mitigating} by large margins and yielding a 36\% and 60\% relative reduction, respectively.
A similar trend holds on AMBER, where C$^2$-DPO achieves a hallucination score of 16.1, lower than both vanilla-DPO (39.9) and C-DPO (17.5).
The improvements generalize to the larger model as well: on LLaVA-v1.5-7B, our method attains a response-level hallucination of 4.8 and a mention-level score of 2.7, along with clear reductions across all AMBER dimensions.
These hallucination reductions come without sacrificing general multimodal reasoning performance, as C$^2$-DPO maintains strong results on ScienceQA, MM-Vet, and TextVQA.
Overall, these results highlight that C$^2$-DPO provides a scalable and model-agnostic method for mitigating object hallucination in MLLMs.

\label{subsec:main_results}

\subsection{Ablations and Further Analysis}
\label{subsec:ablations}
In this section, we provide a comprehensive analysis of C$^2$-DPO through ablations and additional studies.
We first examine the contribution of each loss component, and then show that our contextual preference calibration generalizes beyond DPO to variants such as SimPO~\cite{meng2024simpo} and RDPO~\cite{park2024disentangling}, and even to text-only LLM settings.
We further analyze the learning dynamics of CPG and its behavior under varying context lengths, assess robustness to noisy context, and study the sensitivity to loss coefficients, collectively highlighting the robustness and versatility of our approach.

\input{table/ablation_dpo_variants_and_text_only}

\noindent\textbf{Ablation on loss components.}
We analyze how the two components of the C$^2$-DPO objective, $\mathcal{L}_c(x, x')$ and $\mathcal{L}_\text{DPO}(x')$, affect hallucination on Object HalBench and AMBER. 
As shown in Table~\ref{tab:ablations}, optimizing either component alone leads to weaker performance.
In contrast, jointly optimizing both losses produces a clear synergistic effect, yielding substantial improvements over the original baseline.

\noindent\textbf{Effects on different DPO variants.}
To examine whether the Contextual Preference Calibration approach is specific to DPO or can generalize to various preference optimization schemes, we apply our formulation to two representative alternatives: SimPO~\cite{meng2024simpo} and RDPO~\cite{park2024disentangling}.
As shown in Table~\ref{tab:ablation_dpo_variants}, incorporating our formulation—denoted as C$^2$-SimPO and C$^2$-RDPO—consistently reduces hallucination rates.
These results indicate that contextual preference gain is orthogonal to the underlying preference-learning objective and can be seamlessly integrated with different optimization recipes.
This general applicability suggests our method provides a broadly compatible improvement to preference-based multimodal alignment, rather than being limited to a DPO-specific remedy.

\noindent\textbf{Applicability on text-only LLMs.}
To assess whether our Contextual Preference Calibration (CPC) strategy remains effective in text-only LLM settings, we conduct experiments following the experimental protocol of SimPO~\cite{meng2024simpo}.
Using Qwen2.5-Instruct-1.5B~\cite{wang2024qwen2}, we assess performance on the widely used instruction-following benchmark, AlpacaEval 2~\cite{dubois2024length}.
To adapt CPC to the text-only scenario, we instantiate it following Eq.~\ref{eq:main} by replacing both $x$ and $x'$ using only the query, \ie, $x=q, x'=\varnothing$.
As shown in Table~\ref{tab:text_only}, CPC yields consistent improvements across both DPO and SimPO-based models, highlighting the versatility of our formulation and suggesting that CPC provides a unified and effective strategy for improving reliability in preference optimization.

\input{figure/fig_cpg_curve_fig_avg_cpg}

\noindent\textbf{Learning dynamics of contextual preference gain.}
Figure~\ref{fig:cpg_curve} illustrates how Contextual Preference Gain (CPG) evolves during training for C$^2$-DPO compared to the baseline, C-DPO.
C-DPO exhibits near-zero or even negative CPG throughout training, indicating that it fails to leverage contextual signals.
In contrast, C$^2$-DPO shows a clear and steady increase in CPG.
This pattern suggests that C$^2$-DPO becomes progressively more sensitive to input context during preference optimization, which in turn enables the model to better align its choices with the provided context and leads to reduced object hallucination.

\noindent\textbf{Contextual preference gain according to the context lengths.}
Figure~\ref{fig:avg_cpg} illustrates how Contextual Preference Gain (CPG) evolves as the amount of non-hallucinated contextual information increases.
The top figure reports the average CPG as a function of the number of sentences added to the input.
We observe a clear upward trend: CPG steadily increases with more sentences, indicating that richer contextual grounding amplifies the preference difference between grounded and hallucinated responses.
The bottom figure provides a complementary analysis at the word level, showing that CPG also grows consistently as additional words are incorporated.
Together, these results demonstrate that stronger contextual signals yield greater preference separation, highlighting the sensitivity of CPG to the granularity and richness of added context.

\noindent\textbf{Robustness to noisy context.}
We further evaluate robustness when the auxiliary context is noisy.
During training, we randomly mask the auxiliary captions $c$ with probability $p \in \{0.0, 0.3, 0.5\}$.
Table~\ref{tab:noise} shows that C$^2$-DPO consistently achieves lower hallucination rates than C-DPO across all masking levels on Object HalBench.
Importantly, the performance gap widens as the masking probability increases, suggesting that contextual preference calibration enables the model to remain robust even when contextual information is partially corrupted.

\noindent\textbf{Sensitivity analysis on loss coefficients.}
Table~\ref{tab:sensitivity} presents the sensitivity analysis for $\lambda_{c}$ and $\lambda_{u}$, which control the strength of $\mathcal{L}_c(x,x')$ and $\mathcal{L}_\text{DPO}(x')$.
The results show that the losses are most effective with weights in the 0.3 to 0.5 range, while 0.1 provides comparatively weaker influence.
This narrow region of strong performance remains consistent across both models, indicating that C$^2$-DPO works reliably without extensive hyperparameter tuning.

\input{table/robustness_and_hp_sensitivity}

\subsection{Qualitative Analysis}
\label{subsec:qual}
Figure~\ref{fig:qual} shows qualitative comparisons between the baseline model and ours.
The baseline exhibits hallucinations (``a hot dog with \textit{mustard}''). 
Although it correctly identifies the object as a hot dog, it mistakenly describes it as having \textit{mustard}, revealing a failure to attend to fine-grained visual details.
In contrast, C$^2$-DPO produces an accurate description (``the hot dog is covered in \textit{chili}''), showing that it inspects such details more carefully.
It also identifies additional cues, such as a ``bottle of water'', which the baseline overlooks.
This reflects our objective, which calibrates preference toward grounded evidence rather than frequent co-occurrences, a common source of hallucination.

Figure~\ref{fig:qual2} further illustrates how each model uses contextual information provided alongside the image.
CPG reveals this directly: the baseline C-DPO shows a low score, indicating that it underutilizes the key cue (\eg, ``a \textit{bus stop} on the left side of the image'').
In contrast, our C$^2$-DPO attains a much higher CPG score, correctly referencing ``the \textit{bus stop}'' and grounding its response in the described surroundings.
Here CPG tracks observable behavior: its higher value coincides with explicit use of the caption, illustrating at the instance level what Section~\ref{subsec:context_blindness} shows in aggregate.
By enabling stronger context-aware reasoning, C$^2$-DPO mitigates hallucinations more faithfully in multimodal understanding.

\input{figure/fig_qual_captioning}
\input{figure/fig_qual_w_cpg}

%% file: table/main_table.tex
\begin{table*}[t!]
    \caption{
        \textbf{Comparison of hallucination mitigation methods in MLLMs.}
        The best and second-best results are highlighted in \textbf{bold} and \underline{underline}, respectively.
        All comparisons are conducted under identical model size and architecture.
        "Rsp." and "Men." denote response-level and mention-level hallucination degrees, while "Hal." and "Cog." represent the Hallucination Score and Cognitive Score, respectively.
    }
    \label{tab:main_table}
    \centering
    \setlength{\tabcolsep}{1pt}
    \begin{adjustbox}{width=\textwidth}
        \begin{tabular}{llcccccc|ccc}
            \toprule
            \multirow{3}{*}{\textbf{Model}} & \multirow{3}{*}{\textbf{Method}} & \multicolumn{6}{c|}{\textbf{Hallucination Benchmarks}} & \multicolumn{3}{c}{\textbf{General Benchmarks}} \\
            & & \multicolumn{2}{c}{\textbf{Object HalBench}} & \multicolumn{3}{c}{\textbf{AMBER}} & \multicolumn{1}{c|}{\textbf{HallusionBench}}
            & \multicolumn{1}{c}{\textbf{ScienceQA}} & \multicolumn{1}{c}{\textbf{MM-Vet}} & \multicolumn{1}{c}{\textbf{TextVQA}} \\
            \cmidrule(lr){3-4} \cmidrule(lr){5-7} \cmidrule(lr){8-8} \cmidrule(lr){9-9} \cmidrule(lr){10-10} \cmidrule(lr){11-11}
             & & Rsp. $\downarrow$ & Men. $\downarrow$ & CHAIR $\downarrow$ & Hal. $\downarrow$ & Cog. $\downarrow$ & Question Acc. $\uparrow$ & Image Acc. $\uparrow$ & Overall $\uparrow$ & Acc. $\uparrow$ \\
            \midrule
            \multirow{12}{*}{\textbf{LLaVA-v1.5-7B}~\cite{liu2024improved}}
            & Base                                          & 52.7 & 28.0 & 8.4 & 35.5 & 4.0 & 46.86  & 66.8 & 31.0 & \textbf{58.2} \\
            & VCD~\cite{leng2024mitigating}                 & 51.3 & 25.9 & 9.1 & 39.8 & 4.2 & --     & 68.7 & 29.8 & 56.1 \\
            & OPERA~\cite{huang2024opera}                   & 45.3 & 22.9 & 6.5 & 28.5 & 3.1 & --     & 68.2 & 30.3 & \textbf{58.2} \\
            & DoLa~\cite{chuang2024dola}                    & 44.0 & 25.1 & 6.2 & 27.7 & 2.9 & --     & 67.5 & 30.8 & 56.6 \\
            & HA-DPO~\cite{zhao2023beyond}                  & 37.0 & 20.9 & 6.7 & 30.9 & 3.3 & \underline{47.74}  & \textbf{69.7} & 30.6 & \underline{56.7} \\
            & POVID~\cite{zhou2024aligning}                 & 33.4 & 16.6 & 5.3 & 28.7 & 3.0 & 46.59  & 68.8 & 31.8 & 56.6 \\
            & CLIP-DPO~\cite{ouali2024clip}                 & --   & --   & 3.7 & 16.6 & 1.3 & --     & 67.6 & --   & 56.4 \\
            & RLAIF-V~\cite{yu2025rlaif}                    & 7.8  & 4.2  & \textbf{2.8} & 15.7 & \textbf{0.9} & 35.43 & 68.2 & 29.9 & 55.1 \\
            & TPO~\cite{he2024topic}                        & 5.6  & \underline{3.2}  & 3.6 & 20.5 & 1.6 & 40.12  & 67.1 & 25.7 & 55.3 \\
            & vanilla-DPO~\cite{rafailov2023direct}         & 27.4 & 15.9 & 5.4 & 25.5 & 2.5 & \textbf{47.83}  & 69.3 & \underline{32.0} & \textbf{58.2} \\
            & C-DPO~\cite{peng2025mitigating}               & \underline{5.9}  & 3.3  & \underline{3.0} & \underline{14.9} & 1.3 & 47.48  & 69.4 & \textbf{33.4} & \textbf{58.2} \\
            & \textbf{C$^2$-DPO}                            & \textbf{4.8} & \textbf{2.7} & \textbf{2.8} & \textbf{13.8} & \underline{1.2} & 47.03 & \underline{69.5} & \textbf{33.4} & \textbf{58.2} \\
            \midrule
            \multirow{4}{*}{\textbf{Qwen2-VL-Instruct-2B}~\cite{wang2024qwen2}}
            & Base                                          & 16.1 & 8.3  & 6.2 & 35.5 & 2.7 & 51.11 & 76.9 & \textbf{49.9} & \underline{78.2} \\
            & vanilla-DPO~\cite{rafailov2023direct}         & 6.0  & 3.6  & 4.2 & 39.9 & 3.2 & 51.37 & 77.0 & \underline{49.8} & \textbf{78.4} \\
            & C-DPO~\cite{peng2025mitigating}               & \underline{2.5}  & \underline{1.6}  & \underline{2.7} & \underline{17.5} & \textbf{0.8} & \underline{51.55} & \textbf{77.3} & 48.2 & \textbf{78.4} \\
            & \textbf{C$^2$-DPO}                            & \textbf{1.6}  & \textbf{1.0}  & \textbf{2.7} & \textbf{16.1} & \underline{0.9} & \textbf{52.08} & \underline{77.2} & 49.1 & \textbf{78.4} \\
            \bottomrule
        \end{tabular}
    \end{adjustbox}
\end{table*}

%% file: table/ablation_loss.tex
\begin{table*}[t!]
    \caption{
        \textbf{Ablation on loss components.}
        We ablate the contribution of each loss component in C$^2$-DPO.
    }
    \centering
    
    \begin{adjustbox}{width=\linewidth}
        \begin{tabular}{lcccccccc}
            \toprule
            \multirow{2}{*}{\textbf{Model}} & \multicolumn{3}{c}{\textbf{Method}} & \multicolumn{2}{c}{\textbf{Object HalBench}} & \multicolumn{3}{c}{\textbf{AMBER}} \\
            \cmidrule(lr){2-4} \cmidrule(lr){5-6} \cmidrule(lr){7-9}
            & $\mathcal{L}_\text{DPO}(x)$ & $\mathcal{L}_c(x,x')$ & $\mathcal{L}_\text{DPO}(x')$ & Rsp. $\downarrow$ & Men. $\downarrow$ & CHAIR $\downarrow$ & Hal. $\downarrow$ & Cog. $\downarrow$ \\
            \midrule
            \multirow{4}{*}{\textbf{LLaVA-v1.5-7B}}
            & \cmark &        &                              & 5.9  & 3.3 & 3.0 & 14.9 & 1.3 \\
            & \cmark &        & \cmark                       & 7.1  & 4.1 & 3.1 & 15.9 & 1.2 \\
            & \cmark & \cmark &                              & 5.9  & 3.2 & 3.1 & 15.0 & 1.3 \\
            & \cmark & \cmark & \cmark                       & \textbf{4.8}  & \textbf{2.7}  & \textbf{2.8} & \textbf{13.8} & \textbf{1.2} \\
            \midrule
            \multirow{4}{*}{\textbf{Qwen2-VL-Instruct-2B}}
            & \cmark &        &                              & 2.5  & 1.6 & 2.7 & 17.5 & \textbf{0.8} \\
            & \cmark &        & \cmark                       & 3.3  & 2.6 & 2.7 & 22.5 & 1.1 \\
            & \cmark & \cmark &                              & 3.5  & 2.3 & 3.2 & 18.6 & 1.0 \\
            & \cmark & \cmark & \cmark                       & \textbf{1.6}  & \textbf{1.0} & \textbf{2.7} & \textbf{16.1} & 0.9 \\
            \bottomrule
        \end{tabular}
    \end{adjustbox}
    \label{tab:ablations}
\end{table*}

%% file: table/ablation_dpo_variants_and_text_only.tex
\begin{table}[t]
\centering

\begin{minipage}[t]{0.6\linewidth}
    \caption{
    \textbf{Effectiveness of combining the proposed Contextual Preference Calibration with SimPO and RDPO.}
    The results show that Contextual Preference Calibration generalizes well across preference optimization variants effectively.
    }
    \centering
    \begin{adjustbox}{width=\linewidth}
        \begin{tabular}{lccccc}
        \toprule
        \multirow{2}{*}{\textbf{Method}} & \multicolumn{2}{c}{\textbf{Object HalBench}} & \multicolumn{3}{c}{\textbf{AMBER}} \\
        \cmidrule(lr){2-3} \cmidrule(lr){4-6}
        & Rsp. $\downarrow$ & Men. $\downarrow$ & CHAIR $\downarrow$ & Hal. $\downarrow$ & Cog. $\downarrow$ \\
        \midrule
        SimPO & 3.9 & 2.6 & \textbf{2.9} & 17.8 & 1.0 \\
        \textbf{C$^2$-SimPO} & \textbf{3.3} & \textbf{2.0} & 3.0 & \textbf{14.4} & \textbf{0.8} \\
        \midrule
        RDPO & 3.4 & 2.3 & 2.9 & 38.7 & 2.5 \\
        \textbf{C$^2$-RDPO} & \textbf{1.7} & \textbf{1.2} & \textbf{2.5} & \textbf{29.0} & \textbf{1.6} \\
        \bottomrule
        \end{tabular}
        \label{tab:ablation_dpo_variants}
    \end{adjustbox}
\end{minipage}
\hfill
\begin{minipage}[t]{0.35\linewidth}
    \caption{
    \textbf{Performance on AlpacaEval 2 for LLMs.}
    Applying our Contextual Preference Calibration consistently improves both DPO and SimPO.
    }
    \centering
    \setlength{\tabcolsep}{5pt}
    \begin{adjustbox}{width=\linewidth}
        \begin{tabular}{lcc}
        \toprule
        \multirow{2}{*}{\textbf{Method}} & \multicolumn{2}{c}{\textbf{AlpacaEval 2}~\cite{dubois2024length}} \\
        \cmidrule(lr){2-3}
        & LC (\%) $\uparrow$ & WR (\%) $\uparrow$ \\
        \midrule
        DPO & 24.1 & 27.4 \\
        \textbf{C$^2$-DPO} & \textbf{25.9} & \textbf{29.7} \\
        \midrule
        SimPO & 33.5 & 36.5 \\
        \textbf{C$^2$-SimPO} & \textbf{34.1} & \textbf{37.4} \\
        \bottomrule
        \end{tabular}
        \label{tab:text_only}
    \end{adjustbox}
\end{minipage}

\end{table}

%% file: figure/fig_cpg_curve_fig_avg_cpg.tex
\begin{figure}[t]
\centering

\begin{minipage}[t]{0.42\linewidth}
    \vspace{30pt}
    \centering
    \begin{minipage}[t][0cm][t]{\linewidth}
        \centering
        \includegraphics[width=\linewidth]{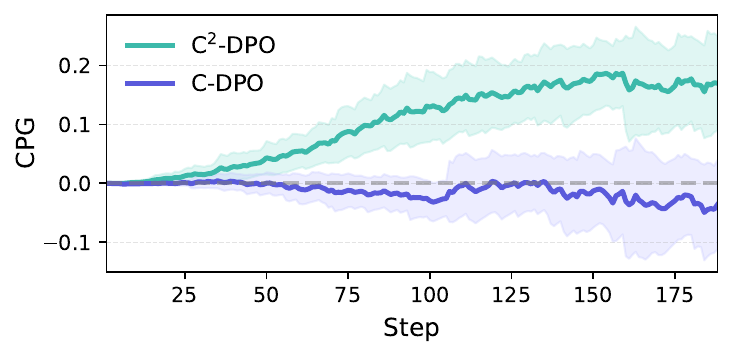}
        \vfill
    \end{minipage}

    \caption{
        \textbf{Contextual Preference Gain (CPG) comparison between C$^2$-DPO and C-DPO.}
        Our C$^2$-DPO steadily increases CPG throughout training, while C-DPO stays near or below zero, indicating its limited ability to utilize contextual signals.
    }
    \label{fig:cpg_curve}
\end{minipage}
\hfill
\begin{minipage}[t]{0.55\linewidth}
    \vspace{0pt}
    \centering
    \begin{minipage}[t][4.9cm][t]{\linewidth}
        \centering
        \includegraphics[width=\linewidth]{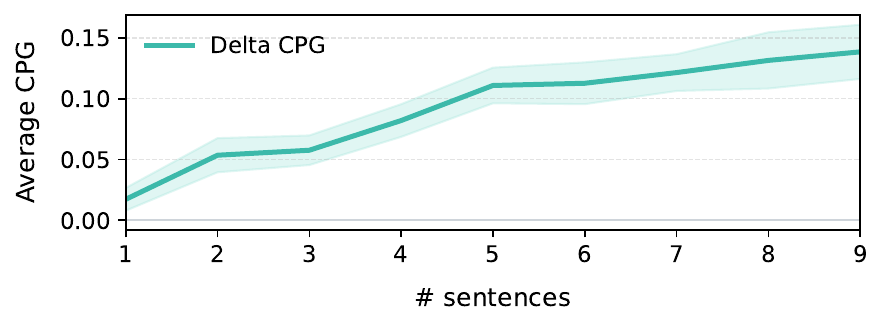}
        \includegraphics[width=\linewidth]{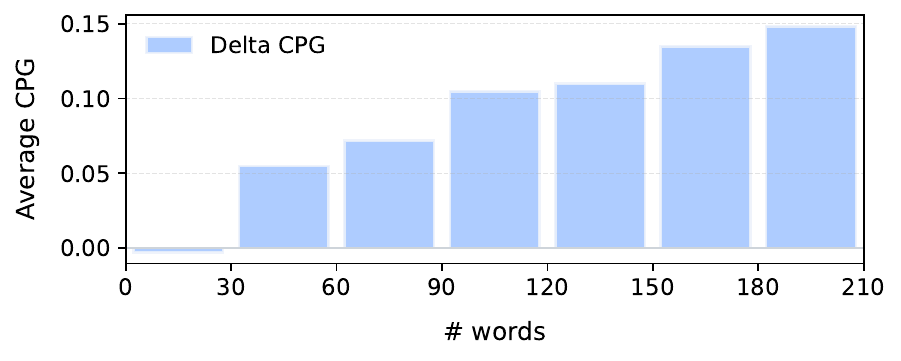}
        \vfill
    \end{minipage}

    \caption{
        \textbf{Average Contextual Preference Gain (CPG) at the sentence (top) and word (bottom) level.}
        CPG increases consistently as more relevant context is added, indicating that richer contextual signals lead to stronger preference separation.
    }
    \label{fig:avg_cpg}
\end{minipage}

\end{figure}

%% file: table/robustness_and_hp_sensitivity.tex
\begin{table}[t]
\centering

\begin{minipage}[c]{0.4\linewidth}
    \caption{
        \textbf{Robustness to context quality.}
    }
    \centering
    \setlength{\tabcolsep}{2pt}
    \begin{adjustbox}{width=\linewidth}
        \begin{tabular}{l c cc}
            \toprule
            \multirow{2}{*}{\textbf{Method}} & \multirow{2}{*}{\textbf{Noise $p$}} & \multicolumn{2}{c}{\textbf{Object HalBench}} \\
            \cmidrule(lr){3-4}
            & & Rsp. $\downarrow$ & Men. $\downarrow$ \\
            \midrule
            C-DPO        & \multirow{2}{*}{0.0} & 2.5  & 1.6  \\
            \textbf{C$^2$-DPO} &                     & \textbf{1.6} & \textbf{1.0} \\
            \midrule
            C-DPO        & \multirow{2}{*}{0.3} & 9.6  & 5.2  \\
            \textbf{C$^2$-DPO} &                     & \textbf{8.5} & \textbf{5.0} \\
            \midrule
            C-DPO        & \multirow{2}{*}{0.5} & 13.0 & 7.2  \\
            \textbf{C$^2$-DPO} &                     & \textbf{8.4} & \textbf{4.6} \\
            \bottomrule
        \end{tabular}
        \label{tab:noise}
    \end{adjustbox}
\end{minipage}
\hfill
\begin{minipage}[c]{0.55\linewidth}
    \caption{
    \textbf{Sensitivity analysis on loss coefficients.}
    }
    \centering
    \setlength{\tabcolsep}{6pt}
    \begin{adjustbox}{width=\linewidth}
        \begin{tabular}{lcccccc}
            \toprule
            \multirow{2}{*}{\textbf{Model}} & \multirow{2}{*}{$\lambda$} & \multirow{2}{*}{\textbf{Value}} & \multicolumn{2}{c}{\textbf{Object HalBench}} \\
            \cmidrule(lr){4-5}
            & & & Rsp. $\downarrow$ & Men. $\downarrow$ \\
            \midrule
            
            \multirow{6}{*}{\textbf{LLaVA-v1.5-7B}}
            & \multirow{3}{*}{\textbf{$\lambda_c$}}
              & 0.1 & 7.4 & 4.3 \\
            & & 0.3 & \textbf{4.8} & \textbf{2.7} \\
            & & 0.5 & 5.5 & 3.1 \\
            \cmidrule(lr){2-5}
            & \multirow{3}{*}{\textbf{$\lambda_u$}}
              & 0.1 & 5.5 & 3.2 \\
            & & 0.3 & 5.6 & 3.1 \\
            & & 0.5 & \textbf{4.8} & \textbf{2.7} \\
            
            \midrule
            
            \multirow{6}{*}{\textbf{Qwen2-VL-Instruct-2B}}
            & \multirow{3}{*}{\textbf{$\lambda_c$}}
              & 0.1 & 4.1 & 2.6 \\
            & & 0.3 & 2.0 & 1.3 \\
            & & 0.5 & \textbf{1.6} & \textbf{1.0} \\
            \cmidrule(lr){2-5}
            & \multirow{3}{*}{\textbf{$\lambda_u$}}
              & 0.1 & 2.4 & 1.5 \\
            & & 0.3 & \textbf{1.6} & \textbf{1.0} \\
            & & 0.5 & 2.0 & 1.3 \\
            
            \bottomrule
        \end{tabular}
    \end{adjustbox}
    \label{tab:sensitivity}
\end{minipage}

\end{table}

%% file: figure/fig_qual_captioning.tex
\begin{figure*}[t]
    \centering
    \includegraphics[width=1.0\textwidth]{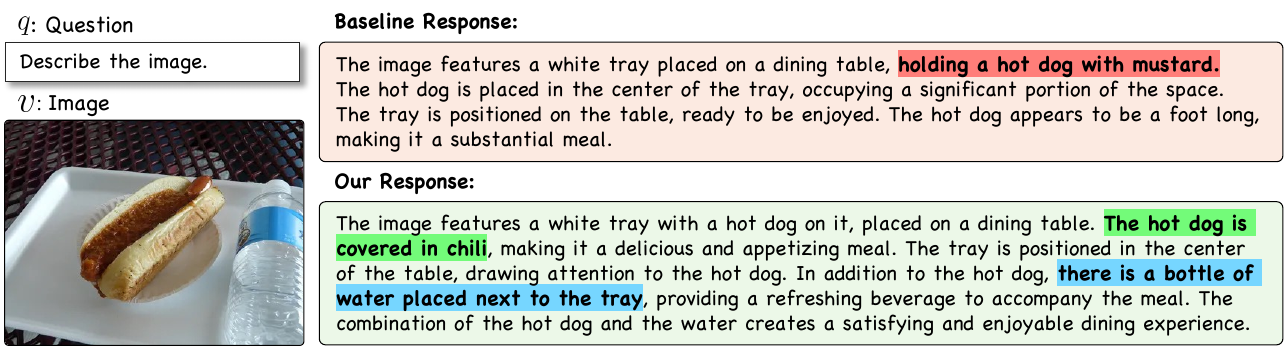}
    \caption{
        \textbf{Qualitative example on detailed captioning.}
        We compare the outputs from models trained with the baseline and our C$^2$-DPO. Red, green, and blue denote hallucinated details, correct details, and additional correct details captured only by our model, respectively.
        Visualization is done on AMBER benchmark with LLaVA-1.5-7B.
    } 
    \label{fig:qual}
\end{figure*}

%% file: figure/fig_qual_w_cpg.tex
\begin{figure}[t!]
    \centering
    \includegraphics[width=1.0\linewidth]{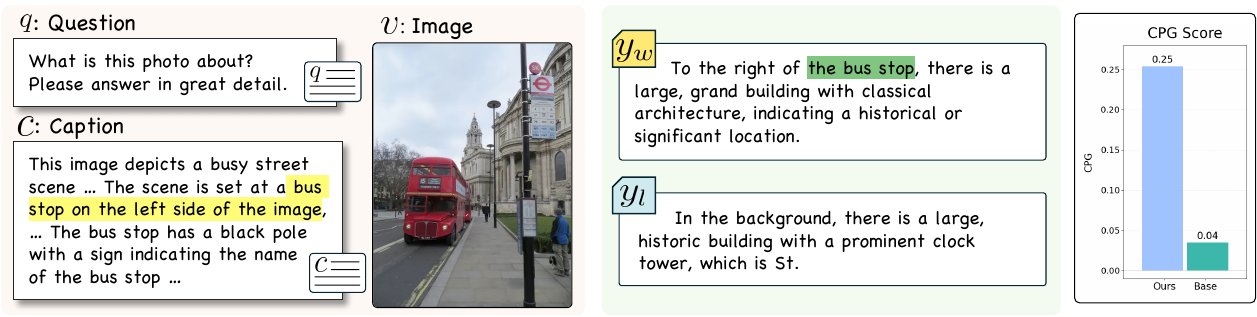}
    \caption{
        \textbf{Qualitative comparison of context utilization between our C$^2$-DPO and the baseline C-DPO.}
        Yellow and green highlights mark context-relevant details from the provided caption that are correctly reflected in the positive response. The bar chart (right) reports the corresponding Contextual Preference Gain (CPG), which is substantially higher for C$^2$-DPO.
    } 
    \label{fig:qual2}
\end{figure}

%% file: sec/6_conclusion.tex
\section{Conclusion}
In this work, we investigated whether existing preference optimization methods truly leverage contextual information for grounding, and found that they do not.
To quantify this, we introduced Contextual Preference Gain (CPG), which revealed two key findings: CPG is strongly and negatively correlated with hallucination rate, and existing DPO-style methods exhibit \emph{context-blind} behavior with CPG concentrated near zero.
Motivated by this, we proposed Context-Calibrated DPO (C$^2$-DPO), which directly maximizes CPG through a simple contrastive regularization while preserving correct preference ordering under degraded context.
Extensive experiments demonstrate that C$^2$-DPO substantially reduces object hallucination without compromising general reasoning ability, and generalizes to other preference optimization variants and text-only settings.
Beyond object hallucination, our results suggest that explicitly modeling how context shapes preference is a broadly useful principle for preference optimization.
As C$^2$-DPO relies only on the contrast between a full and a degraded input, it extends naturally to other modalities and tasks where context can be defined, which we leave for future work.

\noindent\textbf{Acknowledgment.}
This work was supported by the InnoCORE program of the Ministry of Science and ICT (N10250156, 30\%), Institute of Information \& communications Technology Planning \& Evaluation(IITP) grant funded by the Korean government(MSIT) (RS-2025-25442149, LG AI STAR Talent Development Program for Leading Large-Scale Generative AI Models in the Physical AI Domain, 30\%), and the Institute of Information \& communications Technology Planning \& Evaluation (IITP) grant funded by the Korean government(MSIT) (No. RS-2024-00457882, AI Research Lab Project, 40\%).

%% file: sec/X_suppl.tex
\renewcommand{\thesection}{\Alph{section}}
\renewcommand{\thetable}{\Alph{table}}
\renewcommand{\thefigure}{\Alph{figure}}
\setcounter{section}{0}
\setcounter{table}{0}
\setcounter{figure}{0}
\setcounter{page}{1}
\setcounter{equation}{9}

\section*{Appendix}

\section{Gradient Analysis}
\label{sup:gradient_analysis}
In this section, we analyze the gradient of the contextual preference calibration loss $\mathcal{L}_c$ to characterize its training dynamics.
We first show that the gradient factorizes into a calibration direction and an adaptive per-sample weight, and then demonstrate that the resulting update admits an information-theoretic interpretation in terms of the pointwise mutual information (PMI) between a response and the auxiliary context.

\paragraph{Gradient decomposition and training dynamics.}
By the definition of CPG in Eq.~\eqref{eq:cpg}, the calibration loss in Eq.~\eqref{eq:cpc_loss} can be written compactly as $\mathcal{L}_c(x, x') = -\log \sigma\!\left(\mathrm{CPG}(x, x')\right)$.
Differentiating with respect to the model parameters $\theta$ yields
\begin{equation}
    \nabla_\theta \mathcal{L}_c(x, x') = 
    -\underbrace{\sigma\!\left(-\mathrm{CPG}(x, x')\right)}_{\text{adaptive weight}}
    \cdot
    \underbrace{\nabla_\theta \, \mathrm{CPG}(x, x')}_{\text{calibration direction}}.
    \label{eq:grad_Lc}
\end{equation}
The two factors play distinct roles.
The calibration direction $\nabla_\theta \mathrm{CPG}(x, x')$ increases the preference margin under the full context $x$ while decreasing it under the degraded context $x'$, so ascent along this direction directly maximizes CPG.
The scalar weight $\sigma\!\left(-\mathrm{CPG}(x, x')\right) \in (0, 1)$ adjusts the magnitude of the update for each sample.
It stays close to $1$ when the CPG is small or negative, which corresponds to the context-blind regime identified in Section~\ref{subsec:context_blindness}, and becomes small once the CPG grows large.
As a result, $\mathcal{L}_c$ focuses the training signal on samples where the model fails to exploit the auxiliary context, and its influence gradually vanishes as calibration improves.
A similar weighting appears in the standard DPO gradient, where the update is scaled by how strongly the model currently violates the target ordering.

\paragraph{Connection to pointwise mutual information.}
We next show that the calibration direction can be understood in information-theoretic terms.
Let $x = (v, q, c)$ be the full input and $x' = (v, q)$ its context-degraded counterpart.
For a response $y$, the conditional PMI between $y$ and the context $c$ under the policy $\pi_\theta$ is defined as
\begin{equation}
    \mathrm{pmi}_\theta(y; c \mid v, q)
    = \log \frac{\pi_\theta(y, c \mid v, q)}
                {\pi_\theta(y \mid v, q)\, \pi_\theta(c \mid v, q)}
    = \log \frac{\pi_\theta(y \mid x)}{\pi_\theta(y \mid x')},
    \label{eq:pmi_def}
\end{equation}
where the second equality holds because the joint distribution factorizes as $\pi_\theta(y, c \mid v, q) = \pi_\theta(y \mid v, q, c)\, \pi_\theta(c \mid v, q)$, and $\pi_\theta(c \mid v, q)$ denotes the likelihood the model assigns to the context given the image and query.
In other words, the PMI is the log-likelihood ratio of the response with and without the auxiliary context, and it measures how much conditioning on $c$ raises the log-probability of $y$.

For brevity, we write $\Delta_\theta(x)$ for the preference score $\Delta \hat{r}_\theta(x, y_w, y_l)$ in Eq.~\eqref{eq:pref_score}.
Substituting the implicit reward $\hat{r}_\theta(x, y) = \beta \log \frac{\pi_\theta(y \mid x)}{\pi_{\mathrm{ref}}(y \mid x)}$ into $\mathrm{CPG}(x, x') = \Delta_\theta(x) - \Delta_\theta(x')$, grouping the log-probability terms by response, and applying Eq.~\eqref{eq:pmi_def}, we obtain
\begin{equation}
    \mathrm{CPG}(x, x')
    = \beta \underbrace{\big[
        \mathrm{pmi}_\theta(y_w; c \mid v, q)
      - \mathrm{pmi}_\theta(y_l; c \mid v, q)
      \big]}_{\triangleq \, \Delta\mathrm{pmi}_\theta}
      - \beta \, \Delta\mathrm{pmi}_{\mathrm{ref}},
    \label{eq:cpg_pmi}
\end{equation}
where $\Delta\mathrm{pmi}_{\mathrm{ref}}$ is the same PMI gap computed under the reference model.
This term is constant in $\theta$ since $\pi_{\mathrm{ref}}$ is frozen.
Combining Eq.~\eqref{eq:cpg_pmi} with Eq.~\eqref{eq:grad_Lc} then gives
\begin{equation}
    \nabla_\theta \mathcal{L}_c(x, x')
    = -\beta \, \sigma\!\left(-\mathrm{CPG}(x, x')\right)
      \nabla_\theta \big[
        \mathrm{pmi}_\theta(y_w; c \mid v, q)
      - \mathrm{pmi}_\theta(y_l; c \mid v, q)
      \big].
    \label{eq:grad_pmi}
\end{equation}

\paragraph{Interpretation.}
Equation~\eqref{eq:grad_pmi} shows that each gradient step on $\mathcal{L}_c$ raises the PMI between the context and the preferred response while lowering it for the dispreferred one.
Since the per-sample PMI is the pointwise integrand of the conditional mutual information $I_\theta(Y; C \mid V, Q) = \mathbb{E}\!\left[\mathrm{pmi}_\theta(y; c \mid v, q)\right]$, minimizing $\mathcal{L}_c$ amounts to optimizing a contrastive surrogate of this quantity.
Rather than maximizing the expected PMI itself, which would encourage dependence on the context regardless of response quality, the loss widens the PMI gap between $y_w$ and $y_l$ and thus makes the context informative specifically about the preferred response.
Standard DPO behaves differently.
It maximizes a likelihood-ratio margin at a fixed context, so it is agnostic to how much of that margin comes from $c$, and its loss can in fact be driven to zero without the context contributing to the preference at all.
This observation is consistent with the near-zero CPG distributions reported in Section~\ref{subsec:context_blindness}.

\section{Effect of Alternative Context Degradations}
\label{sub:further_analysis}
The main paper constructs the degraded input by removing the auxiliary caption, which corresponds to $x'_{\text{no-cap}} = (v, q, \varnothing)$.
In this section, we examine whether the proposed Contextual Preference Calibration remains effective when the degraded input is constructed differently.
We consider three alternatives.
The first variant $x'_{\text{no-img}} = (\varnothing, q, c)$ removes the image while keeping the caption.
The second variant $x'_{\text{noisy-img}} = (\tilde{v}, q, c)$ replaces the image $v$ with $\tilde{v}$, a copy corrupted by additive Gaussian noise.
The third variant $x'_{\text{rand-cap}} = (v, q, \tilde{c})$ replaces the caption with one randomly drawn from another training sample.
All experiments use Qwen2-VL-Instruct-2B~\cite{wang2024qwen2} as the base model.

Table~\ref{tab:sup_another_context_removal} reports the results.
On Object HalBench~\cite{rohrbach2018object}, every variant clearly outperforms the baseline C-DPO~\cite{peng2025mitigating}, which indicates that the benefit of Contextual Preference Calibration does not hinge on one particular form of degradation.
On AMBER~\cite{wang2023amber}, however, the caption-side variants $x'_{\text{rand-cap}}$ and $x'_{\text{no-cap}}$ yield larger improvements than the image-side variants $x'_{\text{noisy-img}}$ and $x'_{\text{no-img}}$.
We attribute this gap to the asymmetric roles of the two inputs.
The image provides the primary grounding context while the auxiliary caption serves as a secondary one, so degrading the image removes the very signal that responses should be grounded in and weakens the calibration contrast.
Degrading the caption instead isolates the contribution of the auxiliary context and preserves the image as a stable anchor, which explains why $x'_{\text{no-cap}}$ achieves the best overall results and supports our design choice in the main paper.
\input{table/sup_another_context_removal}

\section{Training Details}
\label{sub:training_details}
In this section, we provide the details for reproducing our full training pipeline.

\subsection{Training Dataset}
\label{sup:training_dataset}
We use the SENTINEL preference dataset introduced in C-DPO~\cite{peng2025mitigating}, which provides in-domain, context-aware preference data specifically designed for mitigating object hallucination in MLLMs.
Unlike synthetic or heavily engineered preference collections, SENTINEL constructs discriminative preference data by automatically generating non-hallucinated auxiliary context, which enables the model to distinguish non-hallucinated responses from hallucinated ones.

Each data sample consists of $(v, q, c, y_w, y_l)$, where $v$ is the image, $q$ the query, $c$ an auxiliary non-hallucinated visual description, and $(y_w, y_l)$ denote the preferred and dispreferred responses.
Following prior work, we use the dataset exactly as released without any modification, which ensures a fair comparison with existing methods.
The dataset contains approximately 8.6k preference pairs for LLaVA-v1.5-7B~\cite{liu2024improved} and 12k pairs for Qwen2-VL-Instruct-2B~\cite{wang2024qwen2}.

\subsection{Implementation Details}
\label{sup:implementation_details}
All models are trained for one epoch using LoRA-based preference optimization. 
Table~\ref{sup:hyperparameter_configs} summarizes the full set of optimization and calibration hyperparameters used across all experiments. 
Training is performed on $4 \times$~NVIDIA RTX~4090 GPUs using PyTorch~2.5.1 with CUDA~12.1, and ZeRO stage-2 is enabled for memory-efficient distributed execution.
\input{table/sup_hyperparameter_configurations}

\subsection{Algorithmic Description}
\label{sup:algorithmic_description}
For clarity and reproducibility, we provide the full pseudo-code of our C$^2$-DPO training procedure in Algorithm~\ref{sup:algorithm}. 
The algorithm summarizes how full and degraded contexts are used, how preference scores are computed, and how the combined objective is optimized.

\input{table/sup_algorithm}

\section{Experimental Setup for CPG Analysis}
\label{sup:cpg_analysis}
For the CPG analysis in Section~\ref{subsec:observation}, we train DPO, SimPO, and RDPO~\cite{rafailov2023direct,meng2024simpo,park2024disentangling} on the SENTINEL dataset from C-DPO~\cite{peng2025mitigating}, using the Qwen2-VL-Instruct-2B~\cite{wang2024qwen2} model as the base model.
Each method is trained with its own objective, as summarized in Table~\ref{tab:sup_extensions}.

After training, we compute CPG over 2,000 examples randomly sampled from the training split.
For this evaluation, all preference scores are measured with the DPO implicit reward
\begin{equation}
    \hat{r}_\theta(x, y) = \beta \log \frac{\pi_\theta(y \mid x)}{\pi_{\mathrm{ref}}(y \mid x)},
    \label{eq:dpo_reward_eval}
\end{equation}
regardless of the objective used during training.
This choice provides a single objective-agnostic measure of preference strength, so the reported CPG differences reflect what each method learns rather than how its reward is defined.
\input{table/sup_extensions}

\section{Evaluation Benchmark Details}
\label{sup:evaluation_benchmarks}
We provide a detailed description of the evaluation benchmarks used in our study.

\noindent\textbf{Object HalBench.}
Object HalBench~\cite{rohrbach2018object} is a benchmark designed to measure object hallucination in image-captioning tasks for MLLMs.
It leverages the image captions and segmentation annotations from the COCO 2014 dataset~\cite{lin2014microsoft}.

\noindent\textbf{AMBER.}
AMBER~\cite{wang2023amber} is a multi-dimensional benchmark for evaluating hallucination in MLLMs without relying on an external LLM judge.
The dataset contains 1,004 images paired with task-specific prompt templates that assess three major hallucination types: existence (object presence), attribute(object state), and relational(object-object relations) hallucinations.

\noindent\textbf{HallusionBench.}
HallusionBench~\cite{guan2024hallusionbench} consists of a total of 1129 questions on diverse topics(food, math, geometry, statistics, etc.) and 455 visual-question control pairs while focusing on evaluating fine-grained visual misinterpretation and reasoning errors.
Each sample is annotated with a gold answer and clear hallucination indicator, enabling systematic measurement of whether a model introduces nonexistent objects, misreads attributes, or relies on language priors.

\noindent\textbf{ScienceQA.}
ScienceQA~\cite{lu2022learn} is a multimodal question-answering benchmark designed to evaluate the scientific reasoning capabilities of MLLMs across diverse domains.
The dataset contains over 21,000 multiple-choice questions covering natural science, language science, and social science, with many samples requiring reasoning over diagrams, charts, images, and textual contexts.

\noindent\textbf{MM-Vet.}
MM-Vet~\cite{yu2023mm} contains 200 images and 218 questions to evaluate MLLMs on integrated vision-language capabilities, rather than isolated tasks.
The benchmark defines six core VL capabilities: recognition, OCR, knowledge, spatial awareness, language generation, and math.

\noindent\textbf{TextVQA.}
TextVQA~\cite{singh2019towards} is a visual question-answering benchmark focused on scene-text reasoning, where models are required to read and reason about textual content embedded in images to generate correct answers.
It comprises 45,336 questions across 28,408 images sourced from the OpenImages dataset~\cite{kuznetsova2020open}, filtered to ensure the presence of scene text.

\section{Ablation Study Details}
\label{sup:ablation_details}
This section provides additional details for the ablation experiments reported in Table~\ref{tab:ablation_dpo_variants} and Table~\ref{tab:text_only}.

\noindent\textbf{Extension to other preference optimization methods.}
In applying Contextual Preference Calibration to other preference optimization methods, no additional modification is required beyond replacing the components of the original C$^2$-DPO formulation.
Concretely, we take the C$^2$-DPO objective in Eq.~(9) and substitute
(i) $\mathcal{L}_{\text{DPO}}$ with each method’s PO objective, and
(ii) the implicit reward used inside $\mathcal{L}_c$ with the corresponding formulation summarized in Table~\ref{tab:sup_extensions}.

\noindent\textbf{Applicability to text-only LLMs.}
We apply Contextual Preference Calibration to text-only preference optimization by defining the full and degraded contexts as $x = q$ and $x' = \varnothing$.
Following the C$^2$-DPO formulation in Eq.~(9), we substitute $\mathcal{L}_\text{DPO}$ with each method’s native preference optimization loss and compute $\mathcal{L}_c$ using that method’s implicit reward, as listed in Table~\ref{tab:sup_extensions}.
No further modification is required.
Training follows the SimPO~\cite{meng2024simpo} protocol, where UltraFeedback~\cite{cui2310ultrafeedback} is reannotated using ArmoRM~\cite{wang2024interpretable}.
We use Qwen2.5-Instruct-1.5B~\cite{wang2024qwen2} as the policy model.
Evaluation is conducted on AlpacaEval~2~\cite{dubois2024length} using GPT-4o-mini~\cite{openai2024gpt4omini} as the evaluator, and we report raw win rate (WR) and the length-controlled win rate (LC).

\section{Additional Qualitative Examples}
\label{sup:additional_qual}
This section provides additional qualitative examples that complement the analyses in Figure~\ref{fig:qual} and Figure~\ref{fig:qual2} of the main paper.
Figure~\ref{fig:qual3} expands upon the hallucination mitigation comparison shown in Figure~\ref{fig:qual}, demonstrating how our method improves grounding and reduces both object-level and attribute-level hallucinations across diverse scenarios.
Figure~\ref{fig:sup_qual1} provides extended cases corresponding to the context-utilization analysis in Figure~\ref{fig:qual2}, illustrating how changes in CPG are reflected in the model’s qualitative behavior when auxiliary descriptions are added or removed.

\input{figure/fig_qual3}
\input{figure/fig_sup_qual1}

%% file: table/sup_another_context_removal.tex
\begin{table}[h]
    \centering
    \caption{
        \textbf{C$^2$-DPO performance under alternative context removal.}
    }
    \setlength{\tabcolsep}{2.0pt}
        \begin{tabular}{lccccc}
            \toprule
            \multirow{2}{*}{\textbf{Method}} & \multicolumn{2}{c}{\textbf{Object HalBench}} & \multicolumn{1}{c}{\textbf{AMBER}} \\
            \cmidrule(lr){2-3} \cmidrule(lr){4-4}
            & Rsp. $\downarrow$ & Men. $\downarrow$ & Hal. $\downarrow$ \\
            \midrule
            Base                               & 16.1 & 8.3 & 35.5 \\
            vanilla-DPO                        & 6.0  & 3.6 & 39.9 \\
            C-DPO                              & \underline{2.5} & 1.6 & \underline{17.5} \\
            \midrule
            C$^2$-DPO w/ $x'_\text{noisy-img}$ & \underline{1.7} & \underline{1.1} & 18.1 \\
            C$^2$-DPO w/ $x'_\text{no-img}$    & \textbf{1.6} & \underline{1.1} & 18.2 \\
            C$^2$-DPO w/ $x'_\text{rand-cap}$  & \textbf{1.6} & \textbf{1.0} & \underline{17.1} \\
            C$^2$-DPO w/ $x'_\text{no-cap}$    & \textbf{1.6}  & \textbf{1.0} & \textbf{16.1} \\
            \bottomrule
        \end{tabular}
    \label{tab:sup_another_context_removal}
\end{table}

%% file: table/sup_hyperparameter_configurations.tex
\begin{table}[h]
    \centering
    \small
    \setlength{\tabcolsep}{8.0pt}
    \caption{
        \textbf{Training hyperparameters used in our experiments.}
    }
        \begin{tabular}{lcc}
            \toprule
                                           & \textbf{LLaVA-v1.5-7B} & \textbf{Qwen2-VL-Instruct-2B} \\
            \midrule
                Global batch size          & \multicolumn{2}{c}{64} \\
                Learning rate              & \multicolumn{2}{c}{2e-6} \\
                Learning rate scheduler    & \multicolumn{2}{c}{cosine} \\
                Optimizer                  & \multicolumn{2}{c}{AdamW} \\
                Weight decay               & \multicolumn{2}{c}{0} \\
                Epochs                     & \multicolumn{2}{c}{1} \\
                Trainable parameters       & \multicolumn{2}{c}{LLM linear layers only} \\
                LoRA $r$                   & \multicolumn{2}{c}{128} \\
                LoRA $\alpha$              & \multicolumn{2}{c}{256} \\
                Memory optimization        & \multicolumn{2}{c}{ZeRO stage 2} \\
            \midrule
                $\beta$                    & \multicolumn{2}{c}{0.1} \\
                $\lambda_c$                & 0.3 & 0.5 \\
                $\lambda_u$                & 0.5 & 0.3 \\
                
            \bottomrule
        \end{tabular}
    \label{sup:hyperparameter_configs}
\end{table}

%% file: table/sup_algorithm.tex
\begin{algorithm*}[h]
    \caption{C$^2$-DPO Training}
    \footnotesize
    \label{sup:algorithm}
    \begin{algorithmic}[1]
        \Require Policy model $\pi_\theta$, reference model $\pi_{\text{ref}}$ (frozen)
        \Require Dataset $\mathcal{D} \!=\! \{(v, q, c, y_w, y_l)\}$

        \Require Hyperparameters $\beta,\; \lambda_c,\; \lambda_u$

        \For{each minibatch from $\mathcal{D}$}

            \State \textbf{Compute preference scores} under $x \!=\! (v, q, c)$ and $x' \!=\! (v, q, \varnothing)$:
            \begin{align*}
                \Delta\hat{r}_\theta(x, y_w, y_l) &\leftarrow
                    \beta\log\frac{\pi_\theta(y_w \mid x)}{\pi_{\text{ref}}(y_w \mid x)}
                    - \beta\log\frac{\pi_\theta(y_l \mid x)}{\pi_{\text{ref}}(y_l \mid x)} \\
                \Delta\hat{r}_\theta(x', y_w, y_l) &\leftarrow
                    \beta\log\frac{\pi_\theta(y_w \mid x')}{\pi_{\text{ref}}(y_w \mid x')}
                    - \beta\log\frac{\pi_\theta(y_l \mid x')}{\pi_{\text{ref}}(y_l \mid x')}
            \end{align*}

            \State \textbf{Compute losses}:
            \begin{align*}
                \mathcal{L}_{\text{DPO}}(x)     &\leftarrow -\log\sigma\!\left(\Delta\hat{r}_\theta(x, y_w, y_l)\right) \\
                \mathcal{L}_{c}(x, x') &\leftarrow -\log\sigma\!\left(\Delta\hat{r}_\theta(x,y_w, y_l) - \Delta\hat{r}_\theta(x', y_w, y_l)\right) \\
                \mathcal{L}_{\text{DPO}}(x')    &\leftarrow -\log\sigma\!\left(\Delta\hat{r}_\theta(x', y_w, y_l)\right)
            \end{align*}

            \State \textbf{Update} $\pi_\theta$ by minimizing the combined objective:
            \[
                \mathcal{L}_\text{C$^2$-DPO}(x, x') = \mathcal{L}_{\text{DPO}}(x)
                            + \lambda_c\,\mathcal{L}_c(x, x')
                            + \lambda_u\,\mathcal{L}_{\text{DPO}}(x')
            \]

        \EndFor
        \State \Return $\pi_\theta$
    \end{algorithmic}
\end{algorithm*}

%% file: table/sup_extensions.tex
\begin{table*}[h]
    \centering
    \small
    \caption{
        \textbf{Preference optimization (PO) and Contextual Preference Calibration (CPC) objectives.}   
    }
    \label{tab:sup_extensions}
    \begin{adjustbox}{width=\linewidth}
        \begin{tabular}{lccc}
            \toprule
            \textbf{Method} & \textbf{Implicit reward} $\hat{r}_\theta(x, y)$ & \textbf{PO Objective} $\mathcal{L}_\text{PO}(x)$ & \textbf{CPC Objective} $\mathcal{L}_c(x, x')$ \\
            \midrule
            
            \textbf{DPO} & $\displaystyle \beta\log\frac{\pi_\theta(y \mid x)}{\pi_\text{ref}(y \mid x)} $
            &
                $\displaystyle
                    -\log\sigma(\hat{r}_\theta(x, y_w) - \hat{r}_\theta(x, y_l))
                $
            & \multirow{6}{*}{
                $\displaystyle 
                    -\log\sigma\bigl(
                        \Delta\hat{r}_\theta(x, y_w, y_l) - \Delta\hat{r}_\theta(x', y_w, y_l)
                    \bigr)
                $
            } \\
            \cmidrule(lr){1-3}
            
            \textbf{SimPO} & $\displaystyle \frac{\beta}{\vert y \vert}\log\pi_\theta(y \mid x) $
            &
                $\displaystyle
                    -\log\sigma(\hat{r}_\theta(x, y_w) - \hat{r}_\theta(x, y_l) - \gamma)
                $
            \\
            \cmidrule(lr){1-3}
            
            \textbf{RDPO} & $\displaystyle \beta\log\frac{\pi_\theta(y \mid x)}{\pi_\text{ref}(y \mid x)} + \alpha\vert y \vert $ 
            & 
                $\displaystyle
                    -\log\sigma(\hat{r}_\theta(x, y_w) - \hat{r}_\theta(x, y_l))
                $
            \\
            \bottomrule
        \end{tabular}
    \end{adjustbox}
\end{table*}

%% file: figure/fig_qual3.tex
\begin{figure*}[t!]
    \centering
    
    \includegraphics[width=1.0\textwidth]{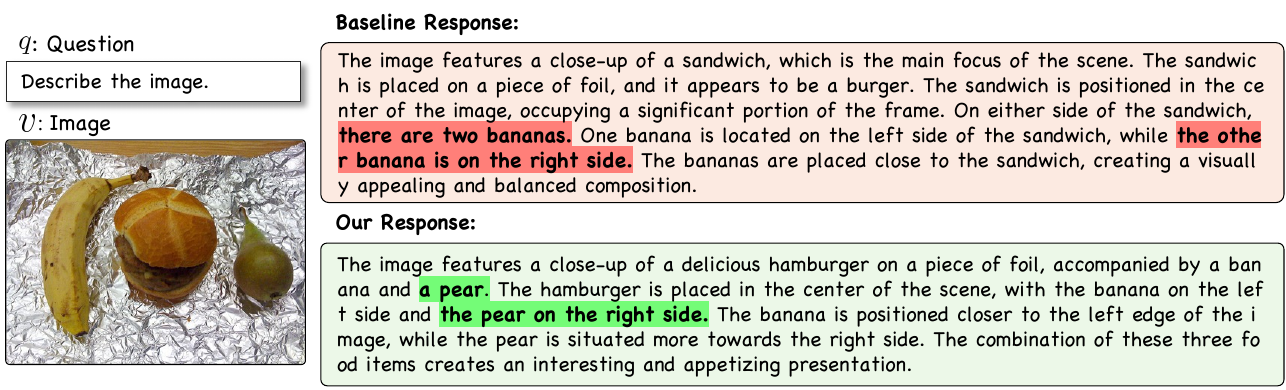}
    \par\vspace{15pt}
    \includegraphics[width=1.0\textwidth]{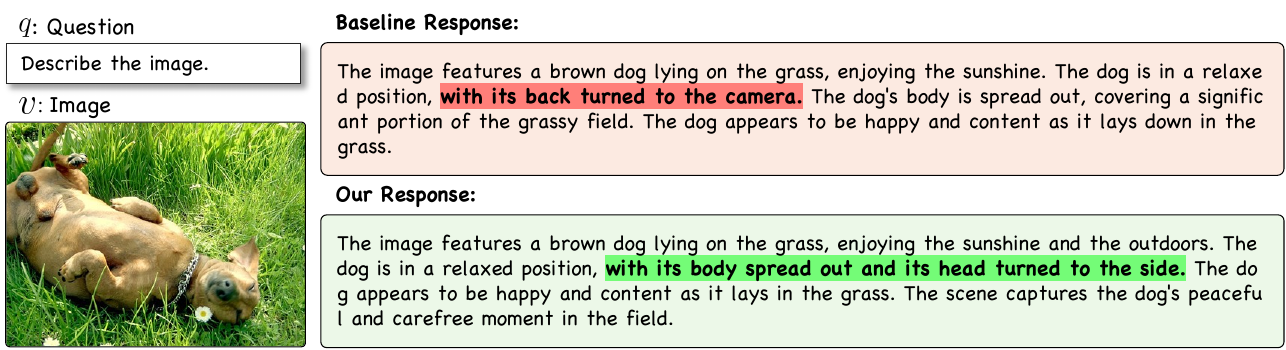}
    \par\vspace{15pt}
    \includegraphics[width=1.0\textwidth]{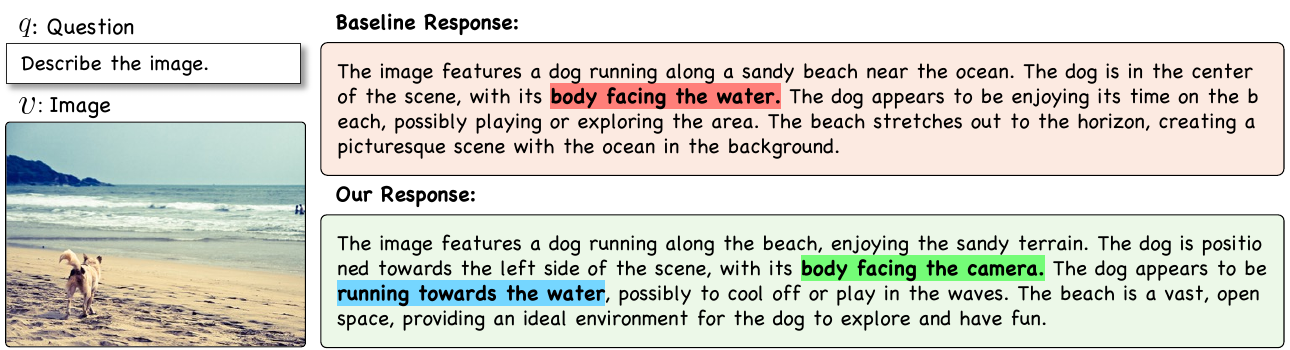}
    \par\vspace{15pt}
    \includegraphics[width=1.0\textwidth]{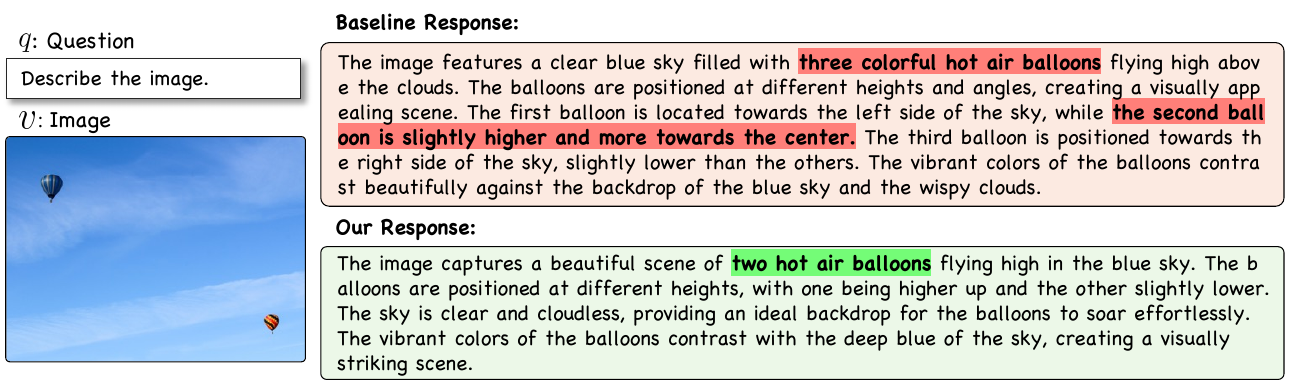}
    
    \caption{
        \textbf{Additional Qualitative examples of image-conditioned description generation.}
    } 
    \label{fig:qual3}
\end{figure*}

%% file: figure/fig_sup_qual1.tex
\begin{figure*}[t!]
    \centering
    
    \includegraphics[width=1.0\textwidth]{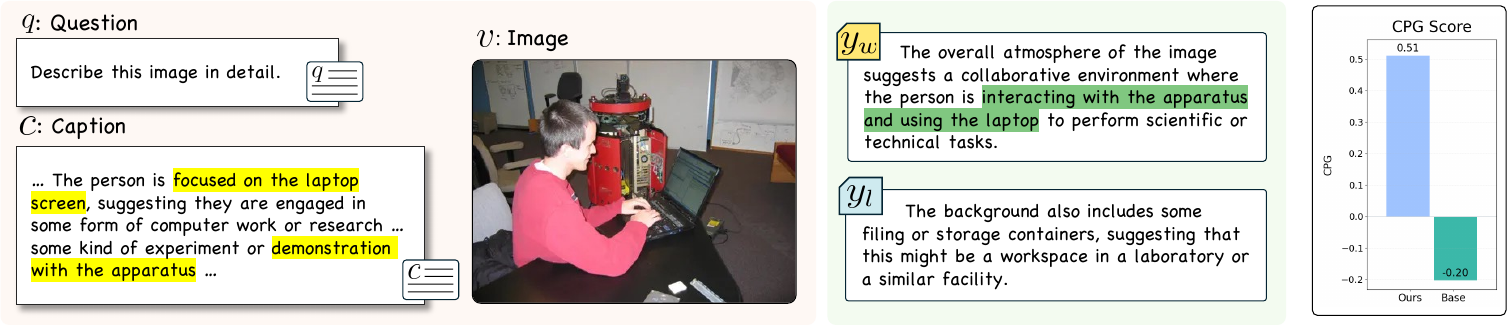}
    \par\vspace{20pt}
    \includegraphics[width=1.0\textwidth]{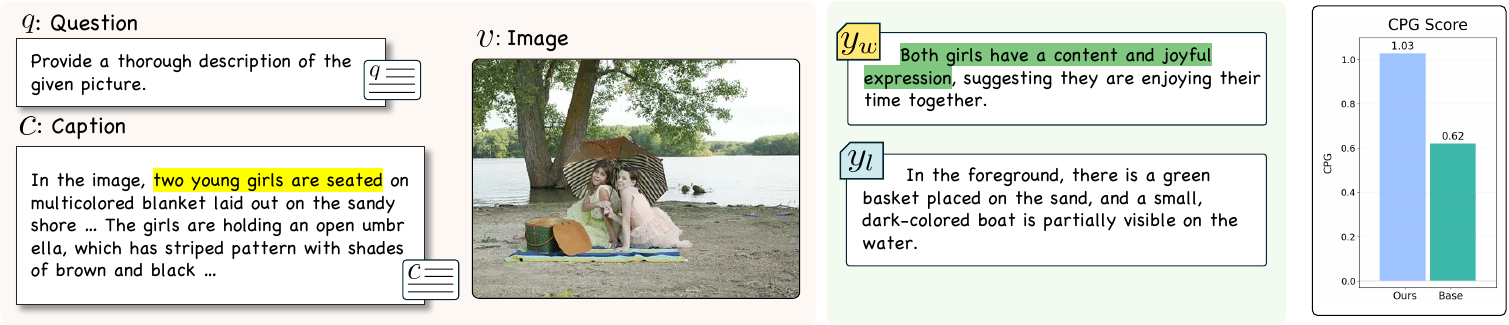}
    \par\vspace{20pt}
    \includegraphics[width=1.0\textwidth]{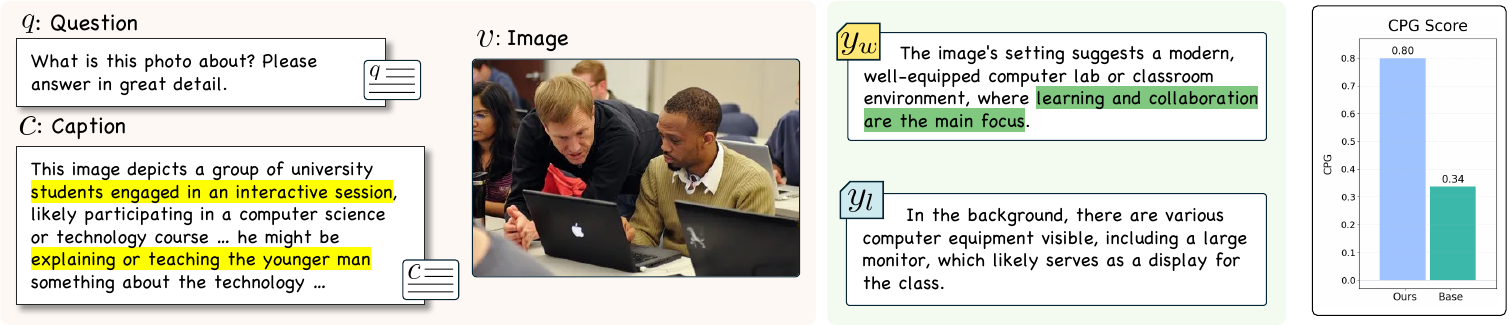}
    
    \caption{
        \textbf{Additional qualitative examples demonstrating how our C$^2$-DPO
                model leverages additional context (caption) more effectively
                than the baseline C-DPO model.
        }
    } 
    \label{fig:sup_qual1}
\end{figure*}